%% file: rldm.tex
\documentclass[11pt]{article} 
\usepackage{rldm,palatino}
\usepackage{graphicx}

\input{math_commands.tex}

\usepackage{wrapfig}
\usepackage{algorithm,algcompatible}

\usepackage[sortcites]{biblatex} 
\title{Revisiting Model-based Value Expansion}

\author{
Daniel Palenicek\\
Intelligent Autonomous Systems\\
Technical University of Darmstadt\\
64289 Darmstadt, Germany \\
\texttt{palenicek@robot-learning.de} \\
\And
Michael Lutter \\
Intelligent Autonomous Systems\\
Technical University of Darmstadt\\
64289 Darmstadt, Germany \\
\texttt{michael@robot-learning.de} \\
\And
Jan Peters \\
Intelligent Autonomous Systems\\
Technical University of Darmstadt\\
64289 Darmstadt, Germany \\
\texttt{jan@robot-learning.de} \\
}

\begin{document}

\maketitle

\begin{abstract}
Model-based value expansion methods promise to improve the quality of value function targets and, thereby, the effectiveness of value function learning.
However, to date, these methods are being outperformed by Dyna-style algorithms with conceptually simpler 1-step value function targets.
This shows that in practice, the theoretical justification of value expansion does not seem to hold.
We provide a thorough empirical study to shed light on the causes of failure of value expansion methods in practice which is believed to be the compounding model error.
By leveraging GPU based physics simulators, we are able to efficiently use the true dynamics for analysis inside the model-based reinforcement learning loop.
Performing extensive comparisons between true and learned dynamics sheds light into this black box.
This paper provides a better understanding of the actual problems in value expansion. We provide future directions of research by empirically testing the maximum theoretical performance of current approaches.
\end{abstract}


\acknowledgements{
This work was funded by the Hessian Ministry of Science and the Arts (HMWK) within the projects ``The Third Wave of Artificial Intelligence - 3AI'' and hessian.AI.
Calculations for this research were conducted on the Lichtenberg high performance computer of the TU Darmstadt.
}

\startmain 

\section{Introduction}

In recent years a large fraction of the reinforcement learning (RL) community has been focused on model-based RL to improve the sample complexity.
Model-based RL algorithms consist of an iterative process of jointly learning a dynamics model from data and then leveraging the learned model in a model-free RL training loop.
The learned models have been used for 
data augmentation~\cite{sutton1990dynaQ,kurutach2018metrpo,janner2019mbpo}, 
improving the value targets~\cite{feinberg2018mve,buckmann2018steve,wang2020dmve,xiao2019adamve}, 
improving the policy gradient~\cite{heess2015svg}
or any combination thereof. These works have proposed various approaches for training the model, new model architectures, (automatically) adapting the rollout horizons and computing better value targets.

A common understanding among most model-based RL approaches is that the compounding model error along modelled trajectories is one of the main problems to be solved or at least avoided.
This compounding model error results in modelled trajectories drifting away from the true trajectory even though they start from the same states and execute the same action sequence. Learning more accurate dynamics models is often believed to be key. 

Two key takeaways that have been used by many of these model-based RL papers and have manifested in recent literature are using
(1) Short model-rollout horizons and
(2) Heteroskedastic ensemble dynamics models. 

\paragraph{Short model-rollout horizons}
As the learned models are at best approximately correct, model errors accumulate with the length of the rollout horizon. Therefore, one common practice introduced by Janner et al.~\cite{janner2019mbpo} is to use shorter rollout horizons with learned models. Otherwise, one exploits the approximation error, and the RL agent fails to learn the task.
This approach can be vaguely thought of as \textit{treating the symptoms} of model errors by behaving pessimistic and cutting rollouts early before the accumulating error can become too large.
In practice, this paradigm is often taken to the extreme by using 1-step model rollouts only.
Furthermore, using shorter rollouts contradicts the theoretical insights we have into value expansion methods.

\paragraph{Heteroskedastic ensemble dynamics models}
Initially proposed by Chua et al.~\cite{chua2018pets}, this model has been widely adopted in most subsequent papers.
The main benefit is that the model learns the aleatoric uncertainty separately from the epistemic uncertainty.
It is an attempt to construct a more capable model architecture which is better suited to represent the environment dynamics. A further benefit is that by explicitly modelling uncertainties they can be used down the line.
\\

Both of these takeaways are an attempt to treat model errors. The first, by reducing the length of the prediction horizon, and the second by explicitly learning uncertainty measures which can be leveraged later down the line.
This leads us to two lines of questioning which might challenge the understanding of the current approaches.

First, if we learned a perfect dynamics model, would this solve all of the problems that current model-augmented actor-critic approaches struggle with?
And if this were the case, could we then just simply use longer horizons and obtain even greater increases in sample efficiency?
The relevance of different possible future research directions is linked directly to the answer of these questions.
If the answer is \textit{yes}, then we should focus future research onto learning more accurate models.
If the answer turns out to be \textit{no}, however, it might be the case that greater accuracy has diminishing returns for model-augmented actor-critic approaches or hinders necessary exploration for example. Striving for more accurate models in these approaches might then not be the most important priority and interesting new research directions could open up. 

Second, are stochastic models really necessary to achieve good results?
Or can deterministic models deliver comparable performance if built and trained carefully?
Current benchmark environments usually feature deterministic dynamics.
Naturally, the question arises, of whether a deterministic model should not be sufficient at learning to model these systems.
If the answer is \textit{yes}, we should revisit deterministic models with the possibility of cutting down complexity.

To come closer to answering these questions, we believe, that there is a need for extensive empirical analysis of the impact of model errors on the training algorithms.
In this paper, we focus on the first line of questioning. We investigate the question of whether learning more accurate dynamics models can still increase the performance of value expansion methods~\cite{feinberg2018mve}.
Therefore, we create an experimental setup with a perfect dynamics model, by replacing the learned dynamics model with an oracle dynamics model.
This allows us to study the theoretical performance of value expansion approaches in isolation without the negative impact of model errors.
Only recently, with the development of GPU based physics simulators, this type of study has becoming computationally feasible. Simulators like BRAX~\cite{brax2021github}, provide GPU accelerated simulation of dynamical systems scaling to thousands of parallel environments.
It allows us to perform fast, oracle dynamics rollouts in the inner model-based RL training loop for entire batches in parallel.
An additional performance benefit is that we are able to perform the training on the GPU only which limits the amount of costly memory transfers from CPU to GPU and back.
%

\section{Maximum-Entropy Model-Based Value Expansion}

We adapt Model-based Value Expansion (MVE)~\cite{wang2020dmve} for the maximum-entropy RL case in order to combine it with a model-free Soft Actor-Critic (SAC)~\cite{haarnoja2018sac} learner.
For this, consider a Markov Decision Process~(MDP)~\cite{puterman2014mdp}, defined by the tuple $\{\mathcal{S}, \mathcal{A}, \mathcal{P}, \mathcal{R}, \rho, \gamma\}$
with state space $\mathcal{S} \subseteq \mathbb{R}^n$ and action space $\mathcal{A} \subseteq \mathbb{R}^m$.
At each time step $t$, the agent observe a state $s_t\in\mathcal{S}$ and samples an action $a_t\in\mathcal{A}$ according to a policy $a_t\sim\pi(\: \cdot\:| \: s_t)$.
The environment returns a next state $s_{t+1}\in\mathcal{S}$ according to the transition probability density function $s_{t+1}\sim\mathcal{P}(\:\cdot\:|\:s_t,a_t)$ and the corresponding scalar reward $r_t = \mathcal{R}(s_t,a_t)$.
The starting state of a trajectory is sampled from the initial state distribution $s_0\sim \rho$. $\gamma$ is a discount factor.
The main objective of maximum-entropy RL is to find a policy $\pi$ that maximizes
$
    J(\pi) = 
    \E_{s_0 \sim \rho} 
    \left\{ \sum_{t=0}^{\infty} \gamma^t (r(s_t, a_t) - \alpha \log \pi(\: \cdot \: | \: s_t)) \right\}
$
with the initial state distribution $\rho$. The actor loss is defined as
$
    J_{\pi}(s_t, a_t) = \alpha \log\big(\pi(a_t | s_t)\big) - Q(s_t, a_t)
$
with
$
    a_t \sim \pi(s_t).
$
In the maximum-entropy case, the value expansion used within the critic loss is described by
\begin{equation*}
    V^{H}(s_0) = \sum_{t=0}^{H-1} \gamma^{t} \Big[ r(s_t, a_t) - \alpha \log \pi(\: \cdot \: | \: s_t)\Big] + \gamma^{H} \Big[ Q(s_H, a_H)  - \alpha \log \pi(\: \cdot \: | \: s_H) \Big].
\end{equation*}
The corresponding critic loss is defined as
$
    J_{Q}(s_{t}, a_{t}, s_{t+1}) = \frac{1}{2} [ Q^{H}_{tar} - Q(s_t, a_t) ]^2
$
with
$
    Q^{H}_{tar}(s_t, a_t, s_{t+1}) = r(s_t, a_t) + \gamma V^{H}(s_{t+1}).
$
We define a learned dynamics model as an ensemble of $N$ probabilistic neural networks
$\hat{\mathcal{P}}_{\Phi}=\{ p_{\phi}^i (s_{t+1},r_t|s_t,a_t)\}_{i=0}^{N}$,
which output mean and variance of the state transition and reward, similar to~\cite{chua2018pets}. 
At inference time, one network in the ensemble is sampled uniformly to capture epistemic uncertainty.

%

\section{Experiments}

In this section, we compare the theoretical performance of MVE with its practical performance.
For this purpose, we construct a version of MVE where we replace the learned dynamics model with an oracle dynamics model. For clarity, we will refer to the latter as \textit{Oracle-based Value Expansion~(OVE)}.
OVE creates a well-defined, artificial environment for studying training performance by eliminating the negative impact of model errors. It lets us answer whether there is still room for performance gains by learning more accurate dynamics models and if, in the absence of model errors, ever longer rollout horizons can increase performance of value expansion methods further.
Our experiments focus on five standard RL benchmark environments: InvertedPendulum, Cartpole Swingup, Hopper, Walker2d and HalfCheetah.
As an efficient GPU based physics simulator, we use BRAX~\cite{brax2021github}, which provides implementations of these benchmark environments.
Our experiments are implemented in JAX~\cite{jax2018github} to integrate seamlessly with BRAX and take full advantage of the GPU.

\subsection{Training Performance}
\begin{figure}
    \centering
    \includegraphics[width=\linewidth]{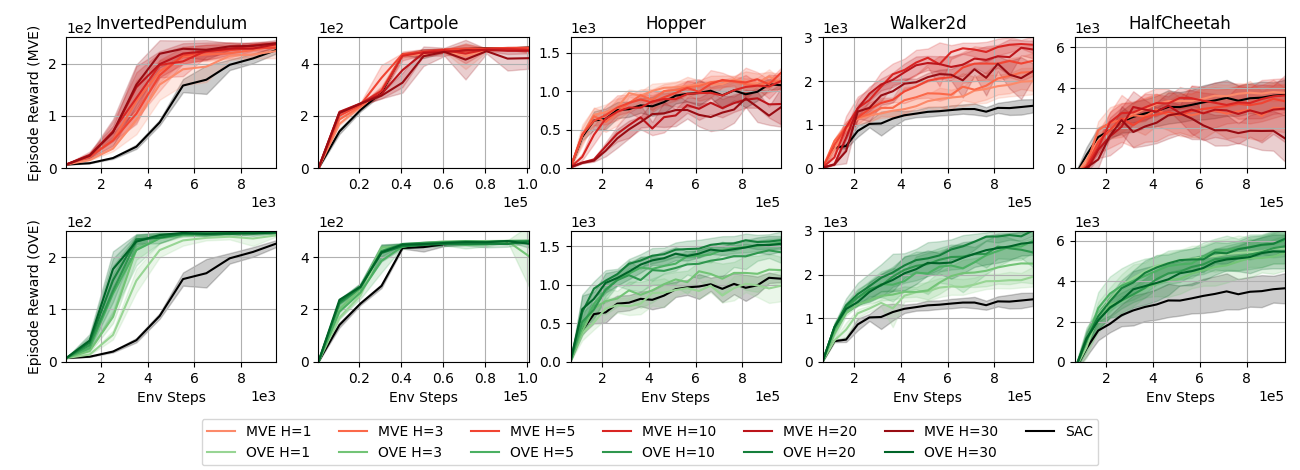}
    \caption{
    MVE training performance (top) and OVE training performance (bottom).
    We evaluate each for multiple rollout horizons $H\in\{1,3,5,10,20,30\}$ and plot the mean and variance across 5 random seeds.
    }
    \label{fig:training_performance}
\end{figure}

We compare the training performance of MVE and OVE. Therefore, we train both algorithms with varying rollout lengths.
Figure~\ref{fig:training_performance} shows the MVE and OVE training performance on the top and bottom row, respectively.
For comparison, we provide a SAC baseline (corresponding to MVE/OVE with a rollout horizon of 0).
We plot the mean and standard deviation across five random seeds.

OVE shows a clear trend that the theoretical improvements of increased horizons can be achieved in the absence of model errors. As expected, the improvements have diminishing returns with increased rollout horizons. Where Walker2d and HalfCheetah suffer a slight performance decrease for $H=30$.
From a practical perspective, there seems to be an optimal trade-off between the benefit of a longer rollout horizon and the increase in computational cost.

Overall, MVE results are split.
While longer rollout horizons assist the training performance in InvertedPendulum, Cartpole and Walker2d, the training performance of Hopper and HalfCheetah suffers immensely.
We assume that the learned model is not accurate enough to produce better value targets for the learning agent in these two environments due to the compounding model error.

\subsection{Diminishing Returns of Longer Rollout Horizons}
\begin{figure}
    \centering
    \includegraphics[width=\linewidth]{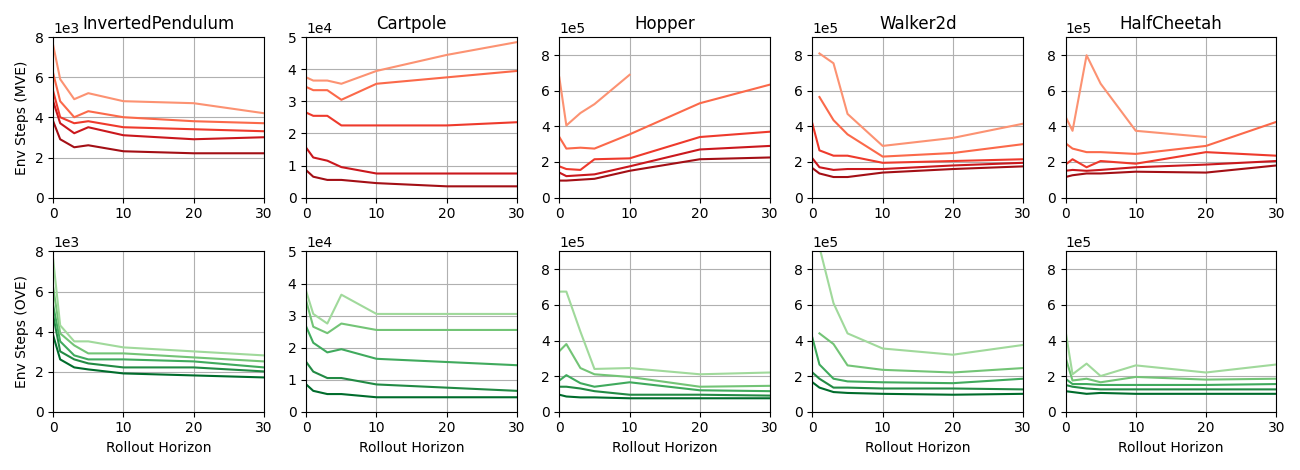}
    \caption{Number of environment steps until MVE/OVE with rollout horizons $H\in\{1,3,5,10,20,30\}$ reach a certain threshold of episode reward. The different thresholds are represented by the different shades of red/green.}
    \label{fig:thresholds}
\end{figure}

We further investigate the diminishing returns in training performance using longer rollout horizons. Figure~\ref{fig:thresholds} shows the number of environment steps that MVE/OVE with a certain rollout horizon requires to \textit{first} reach a set threshold on the episode reward.
The differently shaded lines represent different thresholds. The thresholds are linearly interpolated between a [min, max] episode reward which for the different environments we have picked as follows:
InvertedPendulum = [50, 200],
Cartpole = [100, 400],
Hopper = [250, 1000],
Walker2d = [500, 1900],
HalfCheetah = [500, 3000].

OVE shows the tendency of diminishing improvements for longer rollouts. While most environments show notable improvements by increasing rollout horizons from $H=0$ to $H=5$ or even $H=10$, the lines flatten for horizons $H=\{20,30\}$. Even in the absence of model errors, increasing the rollout horizon appears to reach limitations.

Except for the InvertedPendulum environment, MVE experiments show that in the case of a learned dynamics model, rollout horizons above $H=3$ or $H=5$ indeed hurt the overall performance. Up to a point where it is not able to solve HalfCheetah for $H=30$.
This is clear evidence that more accurate dynamics models could improve MVE training performance for short rollout horizons. However, due to the diminishing returns of increasing rollout horizons it has its practical limitations for longer rollout horizons.
%
%
%

\section{Conclusion}
Our experiments have empirically shown that in the absence of model errors, MVE shows increased performance with longer rollout horizons. Therefore, we conclude that MVE can be made more sample efficient by training more accurate dynamics models. At the same time, we have seen diminishing returns of that improvement with increasing rollout horizons.
Our empirical findings strengthen the theoretical justifications of MVE by Feinberg et al.~\cite{feinberg2018mve} and allow for two streams of future research. First, improving model accuracy through better model training techniques and architectures.
Second, understanding \textit{how} model errors impact value expansion and how the negative impact can be mitigated. This needs more research into analyzing and understanding how these model errors negatively impact training.

In the future, we plan to take a detailed look at the exact nature of the impact of model errors on the learning process and on the generated value targets themselves. We hope that by understanding the effects, we can design more capably algorithms that are more robust to model errors and sample efficient at the same time.

\printbibliography

\end{document}

%% file: math_commands.tex
\usepackage{amsmath,amsfonts,bm}

\def\eqref#1{equation~\ref{#1}}

\def\1{\bm{1}}

\DeclareMathAlphabet{\mathsfit}{\encodingdefault}{\sfdefault}{m}{sl}
\SetMathAlphabet{\mathsfit}{bold}{\encodingdefault}{\sfdefault}{bx}{n}

\newcommand{\E}{\mathbb{E}}

